%% file: sample-1col.tex
\documentclass[a4paper,11pt]{article}

\usepackage[left=2.5cm, right=2.5cm]{geometry}

\usepackage{listings}
\usepackage{xcolor}
\usepackage{graphicx}
\usepackage{amsmath,amsfonts,amssymb}
\allowdisplaybreaks

\usepackage{authblk}
\usepackage{etoolbox,balance}
\usepackage{booktabs,makecell,multirow,array,colortbl,dcolumn,stfloats}
\usepackage{xspace,xstring}
\usepackage[hang,flushmargin]{footmisc}
\usepackage{xcolor}

\usepackage{comment}

\usepackage{biblatex}
\begin{document}

\title{Heterogeneous Encoders Scaling In The Transformer For Neural Machine Translation}

\author{Jia Cheng Hu}
\author{Roberto Cavicchioli}
\author{Giulia Berardinelli}
\author{Alessandro Capotondi}

\affil{University of Modena and Reggio Emilia \\ 
Department of Physics, Informatics and Mathematics \\
name.surname@unimore.it}

\date{}
\maketitle

\input{0_abstract}
\input{1_intro}
\input{2_related_works}
\input{3_method}

\input{4_experimental_setup}

\input{5_experiments}

\input{6_conclusion}

{\sloppy % fix hypenation
\printbibliography}

\end{document}

%% file: 0_abstract.tex
\begin{abstract}
Although the Transformer is currently the best-performing architecture in the homogeneous configuration (self-attention only) in Neural Machine Translation, many State-of-the-Art models in Natural Language Processing are made of a combination of different Deep Learning approaches. However, these models often focus on combining a couple of techniques only and it is unclear why some methods are chosen over others. In this work, we investigate the effectiveness of integrating an increasing number of heterogeneous methods. Based on a simple combination strategy and performance-driven synergy criteria, we designed the Multi-Encoder Transformer, which consists of up to five diverse encoders. Results showcased that our approach can improve the quality of the translation across a variety of languages and dataset sizes and it is particularly effective in low-resource languages where we observed a maximum increase of $7.16$ BLEU compared to the single-encoder model.
\end{abstract}

%% file: 1_intro.tex
\section{Introduction}
Neural Machine Translation (NMT), in Natural Language Processing (NLP), describes the problem of automatically translating a text sentence from one language into another using Neural Networks. Over the past years, translation systems performed increasingly better thanks to the design of more sophisticated and effective models. Models consist of an encoder that extracts a representation from the input sequence in the source language and a decoder that generates the same sentence in the target language. 

There are currently several types of neural processing methods. The most popular ones consist of self-attentive networks (SANs or ANNs) \cite{vaswani2017attention}, recurrent neural networks \cite{kalchbrenner2013recurrent} (RNNs) and convolutional (CNN) ones \cite{zhang2015character, bradbury2016quasi, gehring2017convolutional}. More recently, the works of \cite{lee2021fnet} and \cite{hu2022exploring, hu2022expansionnet} proposed two additional variants. These methods differ in the way they process a sequence: for instance, recurrent neural networks \cite{kalchbrenner2013recurrent} apply the same operation over the sequence in a sequential manner and each element is given access to the previous ones thanks to one or two hidden states. Self-attentive neural networks \cite{vaswani2017attention} instead, make use of the Attention mechanism \cite{bahdanau2014neural, luong2015effective}, a method that constructs representations made of the Softmax-weighted sum of all the elements in the sequence. Each of these approaches presents advantages and drawbacks on its own. For instance, RNNs better capture the sequentiality in the data compared to stateless methods \cite{huang2019attention}, such as CNNs and SANs, that rely on simple numeric encodings. However, as a drawback, the first suffers from computational limitations and a limited receptive field. 

In the presence of a variety of possible processing strategies, State-of-the-Art models often leverage the strength of a combination of multiple approaches in both NLP \cite{chen2018best, hao2019modeling, yang2019convolutional, kim2022squeezeformer, gulati2020conformer} and Vision \cite{hu2022expansionnet, huang2019attention}. The effectiveness of hybrid models outlines the benefits of introducing heterogeneous ways of processing inside the same architecture. However, often in these architectures, only two methods are considered and to the best of our knowledge, little has been done on the combination of a larger number of heterogeneous encoding methods.  
Additionally, different techniques in State-of-the-Art models are often closely intertwined in one single processing block. 
As a matter of fact, there is a lack of consensus on how to combine effectively each technique. This type of integration approach is limiting in terms of applicability since not all encoding strategies can be easily combined. For instance, in a single encoder architecture, combining either the convolution or the self-attention with recurrent methods ultimately results in a recurrent encoder. 

In this work, we propose an investigation into the effectiveness of combining an increasing number of diverse neural network processing methods. In particular, we adopt the Transformer decoder \cite{vaswani2017attention} and an encoder that is made of an increasing number of heterogeneous methods. Moreover,
we propose a very simple encoder combination strategy such as the simple sum. On one hand, it is bound to be less performing than \emph{ad-hoc} hybrid designs. On the other, its simplicity enables the addition of an arbitrary number of independent and diverse encoders. 

The paper is organized as follows. Section~\ref{sec:releated} presents related works and several encoding techniques. Then, in Section~\ref{section_methods}, we introduce the Multi-Encoder Transformer, which encoder combines up to five processing methods: Self-Attention, Convolutional, LSTM, Fourier Transform (FNet), and Static Expansion. In Section~\ref{sec:exp_setup}, we present five translation datasets on which our models are trained and tested. Finally, in Section~\ref{sec:results}, we present the results and discuss the advantages and limitations of our multi-encoder transformer. In particular, we discover that different methods synergize differently when combined with each other and that low-resource languages benefit the most from the increase in the number of heterogeneous encoders.

In summary, our contributions are as follows: \emph{(i)} we analyze the effectiveness of simply summing multiple encoding strategies in the Transformer, in particular adopting the RNN, CNN, SAN, Static Expansion and Fourier Transform (FNet) across a variety of Language translation tasks; \emph{(ii)} we analyze the synergies of the five processing methods and design a dual-encoder, triple encoder, quadruple and quintuple encoder based upon the results; \emph{(iii)} we show that the multi-encoding achieves better performances with respect to the baseline Transformer, despite each encoder presenting a poorer performance in the single instance; \emph{(iv)} we show that low-resource languages are those that benefit the most from combining heterogeneous encoders.

%% file: 2_related_works.tex
\section{Related Works}
\label{sec:releated}

% \textbf{Combining multiple encoders}. 
Many works in several research fields explored the benefits of combining multiple, sometimes complementary, encoding strategies. 
QRNNs \cite{bradbury2016quasi} %  Quasi-recurrent neural networks
tackled the computational burden of RNNs and proposed to exploit the efficiency of the convolution operation while preserving the strong sequential awareness using recurrent pooling layers in various NLP tasks. In Neural Machine Translation
\cite{chen2018best} % combining best of two worlds in NMT
integrated the self-attention layer in the GNMT \cite{wu2016google}, % Google's neural machine translation system: Bridging the gap between human and machine translation
a full RNN network based on stacks of LSTM networks \cite{hochreiter1997long}. Conversely, 
\cite{hao2019modeling} % Modeling recurrence for transformer...  
explored the opposite solution and integrated the LSTM inside the Transformer. The works of \cite{wu2019pay} and
\cite{yang2019convolutional} % Convolutional self-attention networks
merged the self-attention method and convolution on a deep level, forming a moving window in which relationship among sequence tokens are extracted on a local level instead of a global one. On a similar level, in Audio Speech Recognition, best-performing architectures such as the Conformer \cite{gulati2020conformer} % Conformer
and SqueezeFormer \cite{kim2022squeezeformer} % ASR SqueezeFormer
rely on a combination of convolution layers and self-attentive blocks to capture both local and global correlations respectively in the spectrum representation. Similar ideas were applied in other tasks such as Reading Comprehension \cite{yu2018qanet} % Qanet: Combining local convolution with global self-attention for reading comprehension}
and Image Classification \cite{dai2021coatnet}. % Coatnet: Marrying convolution and attention for all data sizes
The adoption of multiple encoders is natural for some problems, such as those involving multiple input sequences \cite{shin2018multi, 
% Multi-encoder transformer network for automatic post-editing
humeau2019poly} % Poly-encoders:architectures and pre-training strategies for fast and accurate multi-sentence scoring
which also led to the investigation of optimal interconnection between the encoders and the decoder \cite{libovicky2018input, %   % Input combination strategies for multi-source transformer decoder
hao2019modeling, chen2018best}. 
Also, the scalability of Transformers has been deeply investigated in a variety of contexts % -> No space 
\cite{wang2022deepnet, jaszczur2021sparse, liu2020very, pham2019very, beltagy2020longformer} which mostly focus on increasing the number of layers or sequence length.

In contrast to these works, our work is the only one that investigates the effectiveness of increasing the number of heterogeneous encoding methods in the Transformer and present hybrid architectures made of an unprecedented number of strategies.

%% file: 3_method.tex
\section{Heterogeneous Multi-Encoder}
%\section{Method}
\label{section_methods}

\subsection{Encoding Strategies}

Given the input sequence $X=\{x_1, x_2, \ldots, x_N\},$ $x_{i} \in \mathbb{R}^{d}$ and $d$ the hidden dimension, we describe five encoding strategies based on recurrency, convolution, self-attention, Fourier transform and static expansion.

\textbf{LSTM.} Recurrent Neural Networks are sequence modelling methods inspired by recurrent functions, which output at time step $t \in \{1,2,\ldots, N\}$ are affected by all the input of the previous steps by means of the hidden states. The hidden state consists of one or multiple vectors depending on the architecture. In this work, we consider the LSTM \cite{hochreiter1997long} encoder whose formula is showcased below:
\[
\left\{
\begin{array}{>{\displaystyle}l@{}}
f_t = sigmoid( W_f x_{t-1} + U_f h_{t-1} + b_f) \\
i_t = sigmoid( W_i x_{t-1} + U_i h_{t-1} + b_i) \\
o_t = sigmoid( W_o x_{t-1} + U_o h_{t-1} + b_o) \\
z_t = tanh(W_c x_{t-1} + U_c h_{t-1} + b_c) \\
c_t = f_t \odot c_{t-1} + i _t \odot z_t \\
h_t = o_t \odot tanh(c_t)
\end{array}
\right.
\]
where $W, U \in \mathbb{R}^{4 \cdot d \times d}$, $b \in \mathbb{R}^{4 \cdot d}$ and $f_t$, $i_t$, $o_t$ are the forget, input, output gates, whereas $h_t$, $c_t$ represent the hidden states. 
We restrict to the uni-directional formulation because the bidirectional processing of the input is already provided by the bidirectional nature of self-attention, which supports the goal of experimenting with diverse encoding strategies.

\textbf{ConvS2S.} Convolution networks were partially introduced in Natural Language Processing to overcome the computational inefficiency of recurrent models \cite{bradbury2016quasi} but later proved to be an effective alternative to other sequence modelling architectures and are even able to achieve State-of-the-Art performances \cite{gehring2017convolutional, yang2019convolutional, liu2020convtransformer, wu2019pay}. In this work, we will focus on the ConvS2S encoder \cite{gehring2017convolutional}. Similar to the Vision case, its working principle lies in the filter banks learning the local and most simple semantics among elements in the sequence during earlier stages and extracting longer and more complex relationships in the higher-order layers. Given an input sequence $X$ and an odd kernel size $k$, the encoder is made of a set of filters $W \in \mathbb{R}^{2 \cdot k \times 2 \cdot d}$ that perform the convolution operation for each time step $t=\{1,2,\ldots, N\}$ in the input sequence. The output dimension is twice the model dimension in order to be fed into the GLU function, which represents the non-linear component of the layer:
\begin{align*}
    GLU(a, b) = a \otimes sigmoid(b)
\end{align*}
where $a, b \in \mathbb{R}^{d}$ are the two halves of the output vector.

\textbf{Self-Attention.} The attention mechanism was first introduced in NLP in \cite{bahdanau2014neural}, and it was applied to recurrent models drastically improving the performances by allowing the encoder information to be distributed on each encoder input rather than one single state vector. The method was later refined in \cite{luong2015effective} and the two works provided the foundation of the Transformer \cite{vaswani2017attention}. Its main component, the self-attentive layer, consists of the following operations (considering the single-head case):   
\begin{align*}
Self\textnormal{-}Attention(Q,K,V) = softmax(\frac{QK^T}{\sqrt{d}})V
\end{align*}
where $Q, K, V \in \mathbb{R}^{N \times d}$ are linear projections of the same input. The method has been proven to be able to extract the syntax and semantics as well as capture long-range relationships of the sequence thanks to the global and bidirectional access to the whole input right at the first stage of the network compared to the previous method.

\textbf{Static Expansion.} The Static Expansion \cite{hu2022expansionnet} proposes to distribute the sequence into an arbitrary number of elements. In its essence, the idea consists of a Forward phase where the input of length is transformed into a new sequence one featuring a new target length, and a Backward where it's transformed back to the original one. In practice, the input is linearly projected four times, producing $K, V_{A}, V_{B}, S \in \mathbb{R}^{N \times d}$. Given the desired length $N_{E}$, $2 \cdot N_{E}$ expansion vectors are considered, denoted as $E_{Q}$, $E_{B} \in \mathbb{R}^{N_{E}\times d}$. The static expansion layer performs three operations.  

\emph{(i).} First, the dot product similarity is computed between the expanded queries and the keys obtaining the length transformation matrix L:
\begin{equation}
    L =  \frac{E_{Q}K^{\intercal}}{\sqrt{d}}
\label{eq:sim_forward}
\end{equation}
The result is fed into a ReLU function and normalized. The sequences featuring new lengths are computed in the following way:
\begin{equation}
\begin{aligned}
P^{fw}_{i} &= \Psi(ReLU((-1)^{i+1}L), \epsilon) \ \ \ i \in {1, 2} \\
F^{fw}_{i} &= P^{fw}_{i}(V_{i} + E_{i}) \ \ \ i \in {1, 2} \\
\end{aligned}
\end{equation}
$\Psi: (X, \epsilon) \rightarrow Y$, $X,Y \in \mathbb{R}^{N_{1} \times N_{2}}$ is the normalization function and $\epsilon \in \mathbb{R}_{> 0}$ the  coefficient ensuring numeric stability. 

\emph{(ii).} Using the same matrix of Equation \ref{eq:sim_forward}, the sequence is transformed back to its original length (Backward step):
        \begin{equation}
        \begin{aligned}
        P^{bw}_{i} &= \Psi(ReLU((-1 )^{i+1} L^{\intercal}), \epsilon) 
        \ \ \ i \in {1, 2}
        \\
        B^{bw}_{i} &= ReLU(P^{bw}_{i}F^{fw}_{i}) 
        \ \ \ i \in {1, 2}
        \\
        \end{aligned}
        \end{equation}

\emph{(iii).} The two results are combined through a sigmoid gate (sigmoid function indicated with $\sigma$):
\begin{equation}
out = \sigma(C) \odot B^{bw}_{1} + (1 - \sigma(C)) \odot B^{bw}_{2}
\end{equation}
It can be seen in the first version of the Static Expansion, the collection of $2 \cdot N_{E}$ expansion vectors are learnable parameters of size $d$. Since the original formulation was developed for Image Captioning instead of NLP tasks, we enrich the formulation by providing each expansion vector more awareness about the sequence by feeding them into a bilinear projection described as:
\begin{equation}
    \begin{aligned}
    E_Q^{'} &= E_Q (\frac{X^{\intercal}(WX)}{\sqrt{T}}) \\
    E_B^{'} &= E_B(\frac{X^{\intercal}(WX)}{\sqrt{T}}) \\
    \end{aligned}
\end{equation}
where $X \in \mathbb{R}^{T \times d}$ is the input sequence, $W \in \mathbb{R}^{d \times d}$ and $E_Q, E_B \in \mathbb{R}^{N_{E} \times d}$ are learnable weights.

\textbf{FNet layer.} The Fourier Transform describes a mathematical operation that converts a function defined in the time domain into its spectral representation. In particular, it extracts the coefficients of its sinusoidal components localized in the frequency domain. The Discrete Fourier Transform consists of the discrete formulation which is designed to address discontinuous and sampled signals. In \cite{lee2021fnet} they adopt such a method and replace the self-attention in BERT, achieving remarkably close performances despite a parameter-less formulation. The discrete Fourier transform is defined as: 
\begin{align*}
    f_{k+1} = \sum_{t=0}^{N-1} x_{t+1} e^{-\frac{2 \pi i}{N}tk}, \ \ 0 \leq k \leq N-1 
\end{align*}
which is denoted in vector form as $\mathcal{F}(X) \in \mathbb{C}^{N \times d}$. The FNet layer is made of two DFTs applied over the sequence dimension first and then over the hidden dimension, finally only the real part of the results is preserved:
\begin{align*}
    \mathcal{F}_{FNet}(X) = Re(\mathcal{F}_{d}(\mathcal{F}_{N}(X)))
\end{align*}
The mixing of tokens along the two dimensions efficiently provides enough accessibility of the sequence to the higher order units, such as non-linear projections, in order to form meaningful compositions of its elements.

\subsection{Multi-Encoder Transformer} 
\label{method_multi_encoder_arch}

\begin{figure*}
    \center\includegraphics[width=0.7\textwidth]{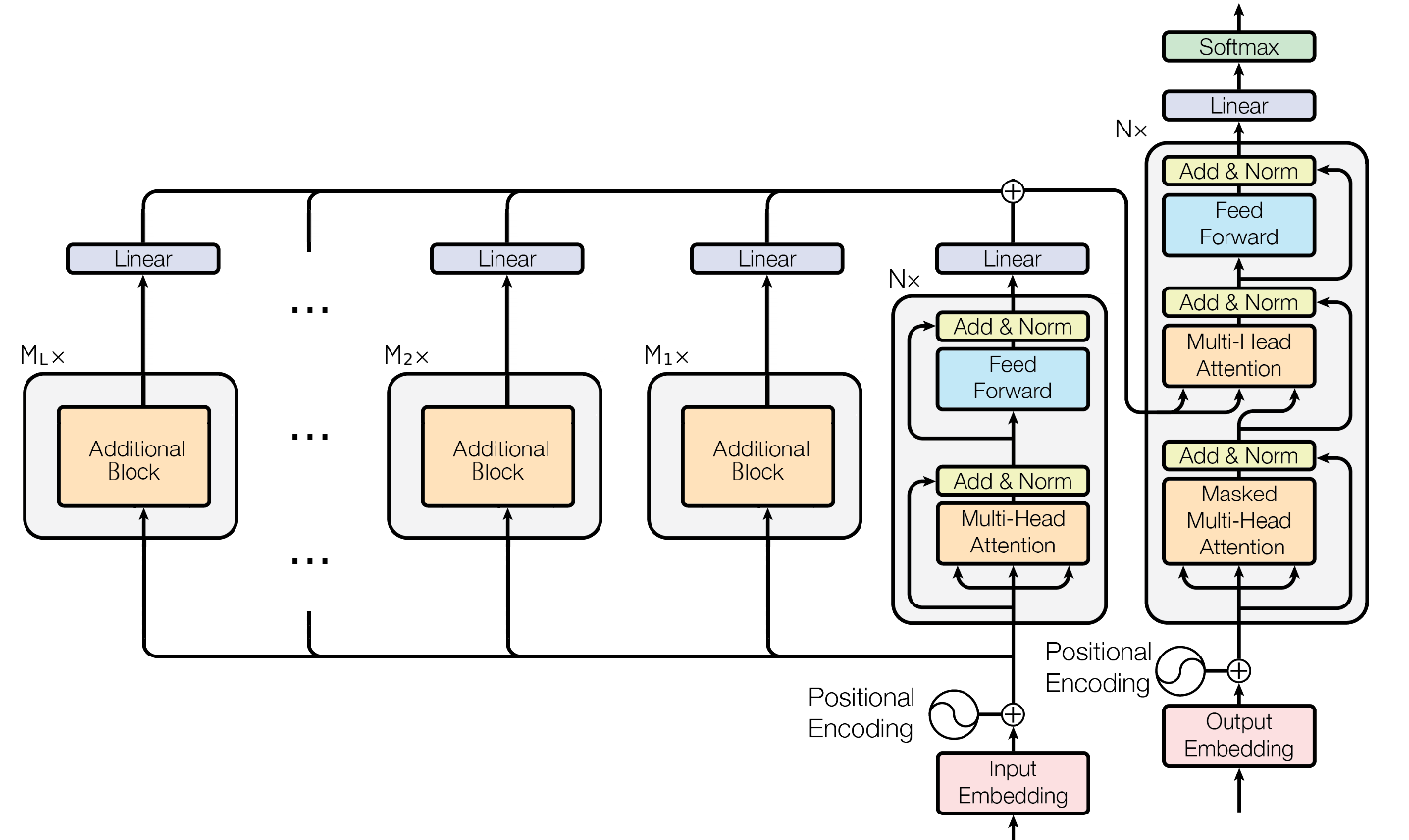}
    \caption{Multi-Encoder Transformer.}
    \label{img_arch}
\end{figure*}

The Multi-Encoder Transformer is made of the Transformer itself and multiple encoding blocks joined together by means of a simple sum or concatenation as depicted in Figure \ref{img_arch}. Encoder blocks, depicted in Figure \ref{img_arch}, consist possibly of a stack of convolution filters, self-attentive layers, LSTMs, Fnet layer, or static expansion. All of them include skip-connections and normalization layers, with the exception of the LSTM for the first and ConvS2S for the latter.

The computational burden of the additional encoder is mitigated by the absence, in almost all cases, of the Feed-Forward layer which represents one of the computational bottlenecks of the transformer model. The only two exceptions are the baseline Transformer encoder, where the component is kept to maintain consistency with the original formulation, and Fnet layers, in which the Feed-Forward is necessary for the otherwise parameter-less definition of the module. 

%% file: 4_experimental_setup.tex
\section{Experimental setup}
\label{sec:exp_setup}

\subsection{Datasets}
\label{sec_datasets}

We used five datasets in our experiments. The IWSLT 2015 English-Vietnamese (En-Vi) corpus, the IWSLT 2017 English-Italian (En-It) corpus, two TED talks in Galician-English (Gl-En), Slovak-English (Sk-En) and a Spain-English (Sp-En) corpus provided by the European Language Resource Coordination (ELRC). We denote the case of ``Gl-En'' and ``Sp-En'' as low-resource language setups. Low-resource languages commonly refer to less studied, re-source scarce, and less commonly taught languages among other definitions \cite{magueresse2020low}. In our work, we denote low-resource languages as those that involve less than 30,000 training pairs.

Sequences whose post-tokenization length is greater than a certain threshold are discarded, which results in a negligible smaller training size but a notably smaller peak GPU memory footprint. In all training instances, the target and the source language vocabulary are shared. Each vocabulary is created using the BPE algorithm \cite{sennrich2015neural}. The maximum number of epochs is chosen based on the number of training steps required for the validation accuracy to start decreasing. For instance, the En-It training requires significantly fewer epochs than En-Vi most likely because of a higher similarity between source and target languages. More Details are reported in Table \ref{tab:data_description}.

\begin{table*}
  \footnotesize
  \centering
  \begin{tabular}{ c c c  c  c  c  c } 
\toprule
 
Source & Language Pair & Total num. pairs &  Max epoch  & Vocab. size & Training max. length \\
 \midrule
 TED Talks & Gl-En & 10017 & 150 &  4139 & 150 \\
 ELRC & Sp-En & 21007 & 200 & 4288 & 200 \\
 TED Talks & Sk-En & 61470 & 350 & 4232 & 150 \\
 IWSLT15 & En-Vi & 133317 & 400 & 4384  & 150 \\
 IWSLT17 & En-It & 231619 & 200 & 4246 & 200 \\
 \bottomrule
 \end{tabular} 
  \caption{
 \label{tab:data_description}
 Data size and training configuration for each translation task. }
\end{table*} 

\subsection{Models}

The baseline model denoted as ``Base" consists of the Transformer with 6 layers, $d$=$512$, $num\_{heads}$=$8$, $d_{ff}$=$2048$. When an encoder is indicated as ``Base", we refer to the ``Self-Attention + FeedForward" case. For the multi-encoder Transformer, the number of additional encoders, as well as the number of additional layers for each block, depends on the experiment configuration and is selected in the following range of values {3, 6, 12, 18}. The Static Expansion layer uses the following 12 choices of static expansion coefficients \textnormal{\{ 6, 6, 12, 8, 12, 8, 6, 6, 12, 8, 12, 8 \}}. The choices of these specific coefficients are arbitrarily made and it is out of the scope of this work to search for the optimal configuration. All kernels in the ConvS2S layers are of width 3. 

\subsection{Training Details}

Words are split according to the Byte Pair Encoding (BPE) \cite{sennrich2015neural}, and no lower casing is applied. Adam optimizer ($\beta_1=0.9, \beta_2=0.98, \epsilon=10^{-9}$) and Label Smoothing regularization (smoothness $\gamma = 0.1$) are adopted.
We use the Noam learning rate scheduler as described in \cite{vaswani2017attention} and 4,000 warm-up steps. Training pairs are sampled without replacement using a target length-oriented bucketing until the mini-batch reaches 4,096 words. 
Finally, models are trained up to the number of epochs reported in Table \ref{tab:data_description}. 

\subsection{Evaluation Method}

We evaluate the performances using the BLEU metric \cite{papineni2002bleu}. In particular SacreBLEU \cite{post2018call}\footnote{SacreBLEU signature: BLEU + case.mixed +
lang.[src-lang]-[dst-lang] + numrefs.1 +
smooth.exp + tok.13a + version.2.0.0}. 
The inference algorithm consists of the Beam Search with beam size 4 and in contrast to the standard practice, no checkpoint averaging is done since the selection criteria are often time-based and lead to different results according to the computational resources.

%% file: 5_experiments.tex
\begin{table}
  \footnotesize
  \centering
  \begin{tabular}{ c c } 
 \toprule
 Encoder & BLEU \\
 \midrule
 Base \cite{provilkov2019bpe} & 31.11 \\
 Base (ours) & 31.13 \\
 LSTM & 29.21 \\
 ConvS2S & 30.95 \\
 Static Exp & 29.40 \\
 FNet & 29.56 \\
 \bottomrule
 \end{tabular}
 \caption{
 \label{tab:single_envi}
 Single encoder performances on the ``En-Vi'' test set. ``Base'' denotes the Self-Attention encoder. }
\end{table}

\begin{table}
  \footnotesize
  \centering
  \begin{tabular}{ c c  c } 
 \toprule
 Task &  N & BLEU \\
 \midrule
 Gl-En & 3 & 13.03 \\
Sp-En & 3 & 27.22 \\
 Sk-En & 6 & 27.15 \\
 En-Vi & 6 & 31.13 \\
 En-It & 6 & 32.86 \\
 \bottomrule
 \end{tabular}

  \caption{
 \label{tab:baselines} 
 Baseline scores on the ``En-Vi'' test set. N represents the number of Transformer layers in the baseline.}
\end{table}

\section{Results}
\label{sec:results}
\subsection{Single Encoder Performances}

We evaluate the performances of each single encoder strategy by replacing the encoder block in the Transformer (with both six encoding and decoding layers) with all five encoding methods in the ``En-Vi'' translation task. This data set is chosen for the analytic experiments throughout the rest of the experiments. 

Each single encoder performance is reported in Table \ref{tab:single_envi}. Since transformers are sensible to hyperparameters and implementation details \cite{nguyen2019transformers, popel2018training} we show that our baseline achieves a very similar performance compared to the ones typically observed in the literature. In the particular case of \cite{provilkov2019bpe} it presents a negligible difference of 0.02 BLEU, paving the way for a fair comparison. In the single-encoder case, the baseline model outperforms all the other ones. Table \ref{tab:baselines} reports the baseline performances across all translation tasks. Hence, the Self-Attention method performs better than the others in the single-encoder configuration.

\subsection{Dual-encoder Transformer and Synergy Study}
\label{sec_ablation}

In Table \ref{tab:ablation_envi} we show how the Dual-Encoder Transformer performs when the baseline, whose encoder is made of Self-Attention layers, is combined with other encoding techniques.
\begin{table}
  \footnotesize
  \centering
  \begin{tabular}{ c c } 
 \toprule
 Encoder & BLEU \\
 \midrule
 Base & 31.13 \\
 Base + ConvS2S\textsubscript{M=6} & 30.34 \\
 Base + ConvS2S\textsubscript{M=12} & 30.21 \\
 Base + FNet\textsubscript{M=6} & 29.20 \\ 
 Base + FNet\textsubscript{M=12} & 29.38 \\ 
 Base + Static Exp\textsubscript{M=6} & 31.51 \\
 Base + Static Exp\textsubscript{M=12} & 31.73 \\
 Base + LSTM\textsubscript{M=6} & 29.52 \\
 Base + LSTM\textsubscript{M=12} & 30.84 \\
 Base + LSTM\textsubscript{M=18} & 31.22 \\
 \bottomrule
 \end{tabular}
 \caption{
 \label{tab:ablation_envi}
 Performances comparison of the dual-encoder Transformer made of the baseline and other encoding methods on the ``En-Vi'' test set.}
\end{table} 
The performance seems to increase proportionally to the number of additional layers $M$ in some cases. This is particularly evident in the case of the LSTM whereas the impact of increasing the number of layers is less significant in other cases. The number of layers for the models presented throughout this work is based on the results in Table \ref{tab:ablation_envi}.

We observe that combining two encoders does not guarantee the overall system performs at least as well as the single encoder case. All dual-encoders yielded a worse score compared to the baseline with the exception of the ``Base + LSTM\textsubscript{M=18}" and ``Base + Static Exp\textsubscript{M=12}" which instead scored better with a margin of 0.09 and 0.60 BLEU. This means that such improvements are caused by neither the architectural structure nor the increase of parameters alone since other instances performed similarly if not worse. This claim is further supported by the poor performances reported in Table \ref{tab:twin_dual_encoder} in which encoders are duplicated and summed together.

\begin{table}
  \footnotesize
  \centering
  \begin{tabular}{ c c c } 
 \toprule 
Method & Single-Encoder & Twin Dual-Encoder ($\delta$) \\
 \midrule
 Base & 31.13 & 30.64 (-0.49) \\
 LSTM & 29.21 & 28.32 (-0.89) \\
 ConvS2S & 30.95 & 30.88 (-0.07) \\
 Static Exp & 29.40 & 29.12 (-0.28) \\
 FNet & 29.56 & 26.51 (-3.05) \\
 \bottomrule
 \end{tabular}
 \caption{
 \label{tab:twin_dual_encoder}
 Twin Dual-Encoder Transformer SacreBLEU scores on the ``En-Vi'' test set. $\delta$ represents the difference w.r.t the baseline.}
\end{table} 

Table \ref{tab:ablation_envi} also shows that, despite the simplicity of the combination strategy, it is possible to design a Dual-Encoder Transformer that performs better than its respective single-encoder counterparts. To further investigate this aspect, given the set of encoding methods E=\{B, L, C, S, F\} representing the self-attention, LSTM, Static Expansion and FNet respectively, we measure the synergy $s_{ij} \in \mathbb{R}$ \ between method $i$ and $j$ simply as the performance difference between the dual-encoder $i+j$ and the single-encoder made of the method $j$. The Synergy matrix $S$ is reported in Table \ref{tab:synergies}. 

\begin{table}
  \footnotesize
  \centering
  \begin{tabular}{  c c  c  c  c  c  c  } 
 % \cline{2-6}
 \toprule
{} & B & L & C & S & F \\
 \midrule
 B & -0.49 & +2.01 & -0.61 & +2.33 & -2.19 \\
 L & +0.09 & -0.89 & -0.18 & +0.19 & -2.08 \\
 C & -0.79 & +1.56 & -0.07 & +0.67 & -1.44 \\
 S & +0.60 & +0.38 & -0.88 & -0.28 & -3.39 \\
 F & -3.76 & -1.73 & -2.83 & -3.23 & -3.05 \\
 \bottomrule
 \end{tabular}
 
  \caption{
 \label{tab:synergies}
 Synergy matrix of dual-encoders on the ``En-Vi'' test set. For brevity, encoders and the number of layers are denoted with letters. B=Base\textsubscript{M=6}, L=LSTM\textsubscript{M=18}, C=ConvS2S\textsubscript{M=6}, S=Static Exp\textsubscript{M=12} and F=FNet\textsubscript{M=6}. Each entry in row $i$ and column $j$ represents the difference in terms of BLEU score between the dual encoder $i+j$ and the single encoder performance of method $j$.}
\end{table} 

Static Expansion and the LSTM combined well with the Transformer encoder, the latter case was expected thanks to the works of \cite{chen2018best, hao2019modeling}.
On the contrary, the ConvS2S case contradicts the effectiveness observed in literature \cite{wu2019pay, yang2019convolutional, gulati2020conformer, kim2022squeezeformer} which highlights the limitations of our trivial combination strategy. The pair ``Base + FNet" seems to perform the worst. 

\subsection{Scaling Encoders}

Based on the Synergy matrix reported in Table \ref{tab:synergies} we increase the number of encoders in the Multi-Encoder Transformer and discuss their performances across all data sets.

To decide which methods are used in the dual and triple encoder configurations, we select the top three values of the formula $max_{i, j \in \{1, \ldots, 5\}}   \{ s_{ij} + s_{ji} \}$ applied on the Synergy matrix reported in Table \ref{tab:synergies}. In particular, the top three values are represented by 2.93, 2.10 and 1.38 corresponding to ``Base + Static Exp\textsubscript{M=12}", ``Base + LSTM\textsubscript{M=18}" and ``LSTM\textsubscript{M=18} + ConvS2S\textsubscript{M=6}". For this reason, we design the Dual-Encoder as ``Base + Static Exp", the Triple-Encoder is made of ``Base + Static Exp + LSTM", the Quadruple-Encoder is made of ``Base + Static Exp + LSTM + ConvS2S" and the Quintuple-Encoder is made of all encoding methods. The number of layers for each additional encoder is configured according to the size of the training set, in particular, for ``Gl-En" and ``Sp-En" we deploy a smaller version of the Multi-Encoder Transformer.

Several observations can be made from the results reported in Table \ref{tab:multi_encoder_results}. \emph{(i).} First of all, the Dual-Encoder outperforms the baseline across all data sets of different sizes and languages, whereas increasing the number of encoders beyond two leads to mixed results. In particular, the performance increase can be appreciated most in the case of two encoders; beyond this case, the improvement is smaller if not even worse. This may explain why State-of-the-Art models in the literature typically combine two deep learning methods at most. \emph{(ii).} The Multi-Encoder Transformer achieved a maximum increase of 5.35 and 7.16 BLEU in the case of ``Gl-En'' and ``Sp-En'' respectively, suggesting that richer encoder representations can compensate for the lack of data. Therefore, adopting multiple heterogeneous encoders can be a suitable strategy for low-resource languages. \emph{(iii).} In the case of bigger datasets, the Multi-Encoder Transformer can be beneficial even though there is no clear relationship between data size and the increase in performance. For example, ``En-It'' and ``Sk-En'' tasks benefit more from the Dual-Encoder configuration compared to ``En-Vi'' despite having a greater and a smaller number of training samples respectively. \emph{(iv).} The case of the Quintuple-Encoder appears to be the breaking point of the benefits of our approach, however, this may be caused by the Fourier Transform method not combining well with others rather than a limitation of the scaling method

Overall, increasing the number of deep learning methods in the encoder can be beneficial but results may depend on the application and data size, and Deep Learning practitioners may consider a trade-off between performances and computational cost in case they decide to follow this strategy. 
    
\begin{table*}
  \footnotesize
  \centering
  \begin{tabular}{ c c c c c c c c c c c } 
 \toprule
 Task & N & Single & M & Dual ($\delta$) & Q & Triple ($\delta$) & U & Quadruple ($\delta$) & V & Quintuple ($\delta$) \\
 \midrule
 Gl-En & 3 & 13.03 & 6 & 17.68 ($+$4.65) & 6 & 17.83 ($+$4.80) & 6 & \underline{18.38} ($+$5.35) & 6 & 12.07 ($-$0.96) \\    
Sp-En & 3 & 27.22 & 6 & 33.63 ($+$6.41) & 6 & 32.81 ($+$5.59) & 6 &  \underline{34.38} ($+$7.16) & 6 & 30.81 ($+$3.59) \\
 Sk-En & 6 & 27.15 & 12 & \underline{27.96} ($+$0.81) & 18 & 27.59 ($+$0.44) & 6 & 26.85 ($-$0.3) & 6 & 26.62 ($-$0.53) \\
 En-Vi & 6 & 31.13 & 12 & \underline{31.73} ($+$0.60) & 18 & 31.71 ($+$0.58) & 6 &  30.19 ($-$0.94) & 6 & 28.90 ($-$2.23) \\
 En-It & 6 & 32.86 & 12 & \underline{33.78} ($+$0.92) & 18 & 33.27 ($+$0.41) & 6 & 32.90 ($+$0.04) & 6 & 32.03 ($-$0.83) \\

 \bottomrule
  \multicolumn{11}{l}{B=Self-Attention, S=Static Expansion, L=LSTM, C=ConvS2S, F=Fourier} \\
 \multicolumn{11}{l}{Single=B, Dual=B+S, Triple=B+S+L, Quadruple=B+S+L+C, Quintuple=B+S+L+C+F}
 \end{tabular}
 \caption{
 \label{tab:multi_encoder_results}
 BLEU scores of the Multi-Encoder Transformer across all translation tasks. N represents the number of Transformer layers. M, Q, U and V represent the number of additional Static Expansion layers, LSTM layers, ConvS2S layers and Fourier layers in the Dual-Encoder, Triple Encoder, Quadruple and Quintuple Encoder respectively. $\delta$ denotes the difference with respect to the baseline. The best scores for each translation task are underlined. }
\end{table*} 

\begin{figure*}
    \center\includegraphics[width=1.0\textwidth]{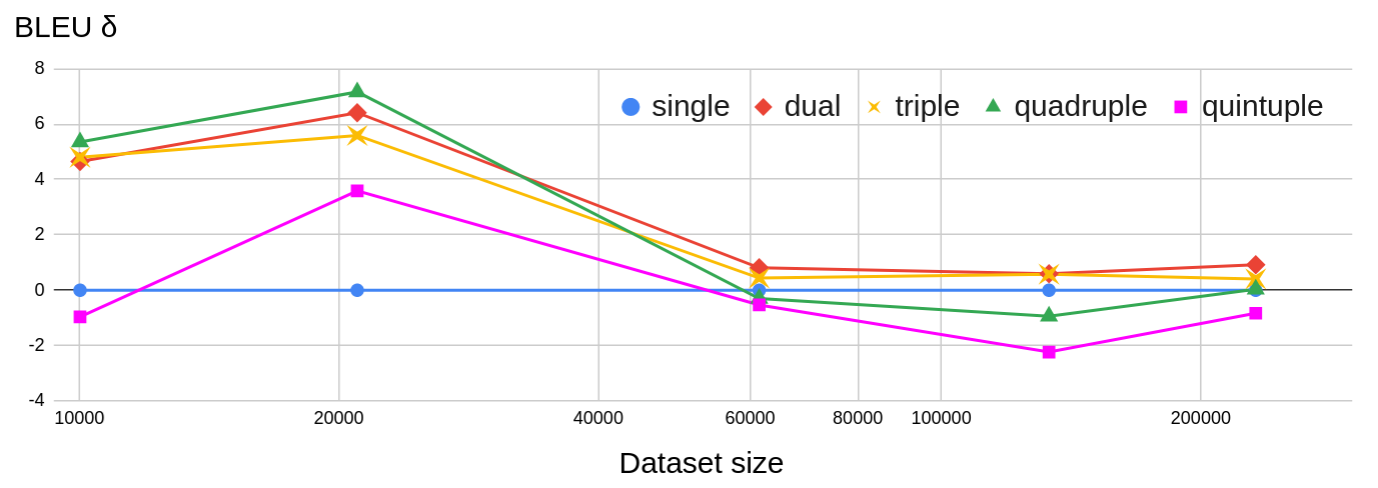}
    \caption{Multi-encoder Transformer performances.}
\label{img_all_performances}
\end{figure*}

\subsection{Memory and Inference Cost}

Although we removed, wherever possible, the feed-forward layer, the additional encoders inevitably introduce a significant additional computational cost and memory footprint as can be seen from Table \ref{tab:cost_limitations}. As a result, the multi-encoder Transformer is increasingly less efficient than the single-encoder baseline depending on the number of additional encoders. Fortunately, benefits can be observed early in the dual-encoder configuration.

\begin{table}
  \footnotesize
  \centering
  \begin{tabular}{ c c c c c c c c c c c } 
 \toprule
 \multirow{2}{*}{Model Task}
  & \multicolumn{2}{c}{Single} & \multicolumn{2}{c}{Dual} & \multicolumn{2}{c}{Triple} & \multicolumn{2}{c}{Quadruple} & \multicolumn{2}{c}{Quintuple} \\
 {} & \multicolumn{1}{c}{$\omega$} & \multicolumn{1}{c}{$\theta$} &  \multicolumn{1}{c}{$\omega$} & \multicolumn{1}{c}{$\theta$} &  \multicolumn{1}{c}{$\omega$} & \multicolumn{1}{c}{$\theta$} &  \multicolumn{1}{c}{$\omega$} & \multicolumn{1}{c}{$\theta$} &  \multicolumn{1}{c}{$\omega$} & \multicolumn{1}{c}{$\theta$}  \\ 
 \midrule
 Gl-En, Sp-En & 5.9 & 26M & 8.3 & 31M  & 11.6 & 46M & 12.8 & 51M & 16.0 & 64M \\
Sk-En, En-Vi, En-It & 11.8 & 48M & 16.7 & 61M & 26.4 & 99M & 27.6 & 109M & 30.8 & 135M \\
 \bottomrule
 \end{tabular}
 \caption{
 \label{tab:cost_limitations}
 Number of parameters (denoted with $\theta$) and the number GFLOPS (denoted as $\omega$) required by the Multi-Encoder Transformer for the forward step of a sequence of 128 tokens in both encoder and decoder.}
\end{table}

\subsection{Limitations of the Study}

The performances reported in this work do not necessarily represent the full capabilities of each model. We focused only on one configuration of the entire hyperparameter space. In particular, we adopted the learning rate scheduler, optimizer, and initialization method of \cite{vaswani2017attention} and the same batch size is used for most configurations. While such choices are ideal for the Transformer, they may not be suitable for all the architectures introduced in this work. The exploration of a variety of configurations is often recommended in order to fully appreciate new models. Such fine-tuning practice is beyond the scope of the work.

%% file: 6_conclusion.tex
\section{Conclusions}

In this work, we conducted an empirical study of the effectiveness of implementing an encoder with an increasing number of heterogeneous Deep learning methods. We considered five Deep Learning methods: self-attention, LSTM, convolution, static expansion, and Fourier transform. We discovered that the Multi-Encoder Transformer can be an effective architectural design for two encoders. However, by increasing the number of encoders beyond two, the impact is limited and can be both positive, such as in the case of low-resource languages, and harmful in other instances. 

Despite combining different encoders with a simple sum, the Multi-Encoder Transformer, particularly the Dual-Encoder configuration, outperformed the baseline across all benchmarks. However, connecting more than two encoders resulted in mixed impact. We suspect this might be due to the extreme simplicity of our combination method. Future works can investigate better combining each encoder representation or even formulating an interconnection module specifically designed to leverage the diversity of different Deep Learning methods.